 \documentclass[pmlr,twocolumn,10pt]{jmlr} 

\usepackage{array}
\usepackage{supertabular}
\usepackage{caption}
\usepackage{multirow}
\usepackage{multicol}
\usepackage{hyperref}
\usepackage{comment}
\usepackage[none]{hyphenat}  
\usepackage{amsmath}
\usepackage{amssymb}

\usepackage[switch]{lineno}

\usepackage{orcidlink}
\usepackage{siunitx}

\theorembodyfont{\upshape}
\theoremheaderfont{\scshape}
\theorempostheader{:}
\theoremsep{\newline}

\jmlrvolume{259}
\jmlryear{2024}
\jmlrsubmitted{LEAVE UNSET}
\jmlrpublished{LEAVE UNSET}
\jmlrworkshop{Machine Learning for Health (ML4H) 2024} 

 \title[Robust Real-Time Mortality Prediction in the ICU using TD Learning]{Robust Real-Time Mortality Prediction in the Intensive Care Unit using Temporal Difference Learning}

\author{
\Name{Thomas Frost} \orcidlink{https://orcid.org/0009-0002-5990-5800} \Email{thomas.frost.21@ucl.ac.uk}\\
\Name{Ken Li} \orcidlink{https://orcid.org/0000-0003-3073-3128} \Email{ken.li@ucl.ac.uk}\\
\Name{Steve Harris} \orcidlink{https://orcid.org/0000-0002-4982-1374} \Email{steve.harris@ucl.ac.uk}\\
\addr Institute of Health Informatics, University College London, United Kingdom
}

\begin{document}

\maketitle

\begin{abstract}
The task of predicting long-term patient outcomes using supervised machine learning is a challenging one, in part because of the high variance of each patient's trajectory, which can result in the model over-fitting to the training data. Temporal difference (TD) learning, a common reinforcement learning technique, may reduce variance by generalising learning to the pattern of \textit{state transitions} rather than terminal outcomes. However, in healthcare this method requires several strong assumptions about patient states, and there appears to be limited literature evaluating the performance of TD learning against traditional supervised learning methods for long-term health outcome prediction tasks. In this study, we define a framework for applying TD learning to real-time irregularly sampled time series data using a Semi-Markov Reward Process. We evaluate the model framework in predicting intensive care mortality and show that TD learning under this framework can result in improved model robustness compared to standard supervised learning methods – and that this robustness is maintained even when validated on external datasets. This approach may offer a more reliable method when learning to predict patient outcomes using high-variance irregular time series data.
\end{abstract}

\paragraph*{Data and Code Availability}
This research makes use of the MIMIC-IV dataset \citep{mimic-1, mimic-2} and the Salzburg Intensive Care dataset \citep{sicdb-1, sicdb-2}, two datasets of de-identified health data collected from Beth Israel Deaconess Medical Center (USA) and University Hospital Salzburg (Austria), respectively. Both datasets are available via the PhysioNet platform \citep{physionet}. Code for this paper is publicly available at \href{https://github.com/tdgfrost/td-icu-mortality}{https://github.com/tdgfrost/td-icu-mortality}.

\paragraph*{Institutional Review Board (IRB)}
For both datasets, the collection of patient information and sharing of de-identified health data for the purposes of research received local ethical approval, as detailed in their respective publications. The research contained herein did not require further independent ethical approval.

\section{Introduction}
\label{sec:intro}
Patients in the Intensive Care Unit (ICU) are amongst the sickest in any hospital and often follow a complex trajectory from admission to discharge (or death). With an average ICU mortality rate between 7-19\%, there is a need for accurate prediction models to help stratify patient risk during admission \citep{icu-mortality-usa, icu-mortality-europe}. 

Traditional approaches rely heavily on scoring systems such as APACHE, SOFA, and SAPS, which provide a mortality risk score from patient features collected at ICU admission. These methods have been externally validated but are cited to have area under the receiver operating characteristic curve (AUROC) scores frequently limited to the 0.70-0.79 range \citep{apache-performance, auroc-curves, sofa-mimic-performance}.

Supervised machine learning algorithms (such as gradient boosted ensembles, artificial neural networks, and Bayesian networks) demonstrate improvement on this baseline, with AUROC scores ranging from 0.80-0.95. However, many of these studies lack high-quality external validation; are limited to one-time predictions at the point of admission (as opposed to ongoing/real-time predictions); and/or limit their mortality time-horizon to the short-term (\(\leq\)72 hours) \citep{ml-icu-mortality-no-ext-1, ml-icu-mortality-no-ext-2, ml-icu-mortality-no-ext-3, ml-icu-mortality-with-ext-1, real-time-icu-mortality-with-ext-2}. Research studies that have validated their models externally typically report a significant deterioration in AUROC performance (10-15\%), consistent with over-fitting to the training data \citep{ml-icu-mortality-with-ext-1, real-time-icu-mortality-with-ext-1}. 

We theorise that the over-fitting of models in these tasks may be related to the significant variance of each patient's trajectory, with diminishing relevance of current features to outcomes far in the future. In reinforcement learning (the branch of machine learning tasked with optimal sequential decision-making), models face a similar challenge of attributing distant rewards under noisy trajectories to present states and actions (the so-called ``credit assignment problem"). A common solution to this has been the use of temporal difference (TD) learning, in which the model is bootstrapped using its own predictions for future states rather than the actual observed (distant) rewards \citep{sutton-barto-book}. This is demonstrated conceptually in \figureref{fig:td-explained}.

\begin{figure}[htpb]
\floatconts
  {fig:td-explained}
  {\caption{Illustrative example of TD learning versus supervised learning. Imagine two patient trajectories (A and B), which share a similar state at \(S_{t+1}\). A model trained using supervised learning might learn to predict a mortality risk of 100\% for \(S^A_t\), and 0\% for \(S^B_t\), based on the observed real outcomes for each state (long red dotted line). A model trained using TD learning would instead learn a 50\% risk for both states, as each inherits the aggregate predicted risk of all future states (short black dotted lines).}}
  {\includegraphics[width=1.0\linewidth]{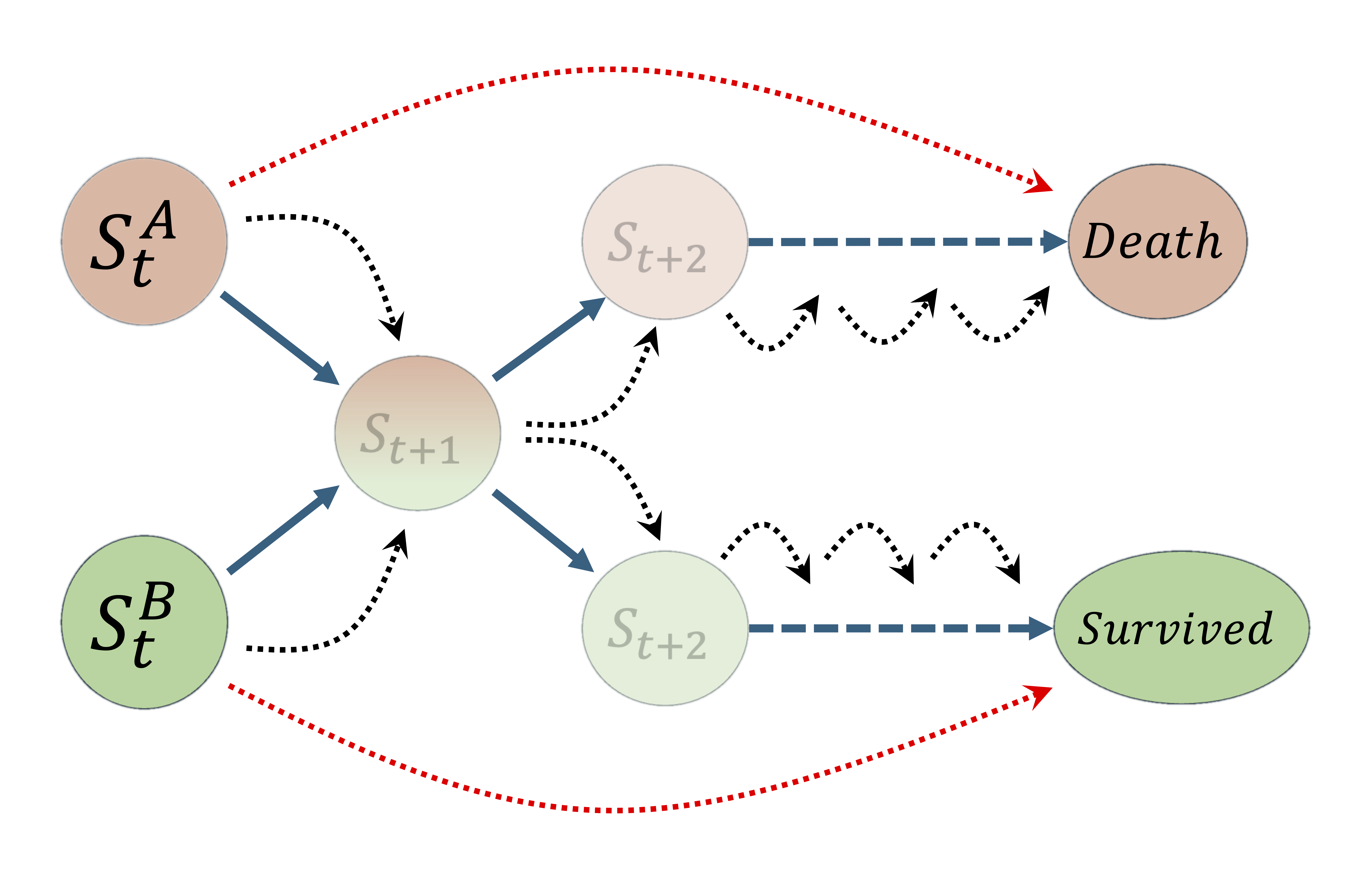}}
\end{figure}

Several reinforcement learning publications already exist exploring the use of TD learning in predicting mortality risk for a given set of actions in intensive care \citep{td-learning-healthcare-1, td-learning-healthcare-2, td-learning-healthcare-3, td-learning-healthcare-4}. However, to date these works focus on using TD learning to generate \textit{counterfactual} predictions based on a \textit{hypothetical} decision policy, making it difficult to evaluate their accuracy without real-world policy deployment \citep{levine2020offline}. Additionally, most such studies artificially aggregate time series data into regular intervals (e.g., every four hours), with only one exception \citep{td-learning-continuous-time} applying TD learning to irregular health data.

In this paper, we generate a set of models using either TD learning or supervised learning for the task of inpatient mortality prediction. The models are trained on more than 65,000 undifferentiated patients using the MIMIC-IV dataset. We describe the mathematical framework for patient states and state transitions (Section \ref{subsection:state-definition} and \ref{subsection:semi-mrp}), the CNN-LSTM base model architecture (Section \ref{subsubsection:model-architecture}), and the training process for each model category (Section \ref{subsubsection:model-training}). We report the AUROC scores for all models on both a hold-out test segment of the MIMIC-IV dataset, as well as an external validation dataset (SICdb) of 21,000 patients (Section \ref{subsection:internal-results} and \ref{subsection:external-results}). We show that models trained with TD learning outperform both supervised learning and clinical score baselines for the task of long-term mortality prediction, and suffer much less overfitting when tested on an external validation dataset. We discuss our interpretations of these results and plans for further work in this area (Section \ref{sec:discuss}).

\section{Methods}
\label{sec:methods}

\subsection{Temporal Difference Learning and Semi-MRP Framework} \label{subsection:semi-mrp}
\subsubsection{Markov Reward Process}
The Markov Reward Process (MRP) describes a Markov chain of states, in which ``memoryless" states transition stochastically to new states over time, and generate rewards \(\mathcal{R}\) in the process (similar to \figureref{fig:td-explained}). It is defined by the tuple \(\{\mathcal{S}, \mathcal{P}_{ss'}, \mathcal{R}, \gamma\}\):

\begin{itemize}
    \item \(\mathcal{S}\): The current state.
    \item \(\mathcal{P}_{ss'}\): The state transition probability matrix from $\mathcal{S}$ to $\mathcal{S'}$, which is independent of any states prior to $\mathcal{S}$.
    \item \(\mathcal{R}\): The immediate reward received.
    \item \(\gamma\): A time discount factor, $\lambda$ $\in$ $[0,1]$.
\end{itemize}

In healthcare, the latent state \(\mathcal{S}_t\) of the patient is not observable, but could theoretically be inferred by the model from observations $\mathcal{O}_t$ recorded over a suitable time period. The transition matrix \(\mathcal{P}_{ss'}\) is thus also unknown, but is expected to be affected by the decision policy of the unobserved clinician.

\subsubsection{TD Learning}
One can train a value function $V(O_t)$ to predict the expected sum of (discounted) rewards \(\mathcal{R}\) collected at all future steps from a set of observations \(\mathcal{O}_t\) as follows \eqref{eq:mc-learning}:

\begin{equation}
\label{eq:mc-learning}
    V(\mathcal{O}_t) \leftarrow \mathbb{E} \left[\sum_{k=0}^{\infty} \gamma^k \mathcal{R}_{t+k+1}\right]
\end{equation}

This equation requires complete episode trajectories using Monte Carlo sampling, which is unbiased but potentially of high variance, particularly as the future trajectory becomes very long. In contrast, via mathematical induction of \eqref{eq:mc-learning}, we arrive at \eqref{eq:td-learning}, the formula for TD learning. This equation uses bootstrapping to reduce the variance by predicting a single step in the future (at the risk of bias).

\begin{equation}
\label{eq:td-learning}
V(\mathcal{O}_t) \leftarrow \mathbb{E} \left[\mathcal{R}_{t+1} + \gamma V(\mathcal{O}_{t+1})\right]
\end{equation}

\subsubsection{Semi-Markov Reward Process}
One limit of MRPs is their dependence on state transitions modelled under regular discrete intervals. To allow for irregularly sampled health data, we instead use a semi-MRP \eqref{eq:smrp-td-learning}, in which transitions are allowed to occur at variable intervals:

\begin{equation}
\label{eq:smrp-td-learning}
V(\mathcal{O}_t) \leftarrow \mathbb{E}_{k \sim \mathcal{D}} \left[\mathcal{\bar{R}}_{t:t+k} + \gamma^k V(\mathcal{O}_{t+k})\right]
\end{equation}

We assume that \(k\) is sampled from a (potentially unknown) stochastic distribution \(\mathcal{D}\), with \(\mathcal{\bar{R}}_{t:t+k}\) representing the discounted sum of rewards captured in the interval between \(t\) and \(t+k\).

\subsubsection{Choice of Reward under SMRP}
We can now convert the formula in \equationref{eq:smrp-td-learning} to a loss function for predicting the terminal outcome (death/discharge) of a patient. We define the following:

\begin{itemize}
    \item The terminal reward is patient mortality, with \(\mathcal{R} = 1\) for death and \(\mathcal{R} = 0\) for survival (i.e., successful discharge).
    \item There are no interim rewards prior to the terminal state.
    \item \(\gamma\) = 1, i.e., the predicted risk of death is the averaged risk from all immediate future states.
\end{itemize}

The above conditions allow us to convert (\ref{eq:smrp-td-learning}) to (\ref{eq:mortality-prediction-td-learning}), producing an averaged mortality risk at time $t$ from 0 to 100\%:

\begin{equation}
\label{eq:mortality-prediction-td-learning}
V(\mathcal{O}_t) \leftarrow \mathbb{E}_{k\sim D} \left[\begin{cases}
\mathcal{R}, & \small \text{if } \mathcal{O}_t \text{ is terminal} \\
V(\mathcal{O}_{t+k}), & \small \text{otherwise}
\end{cases}\right]
\end{equation}

We will now describe the processing steps to accommodate health data into this framework.

\begin{figure*}[t]
    \centering
    \includegraphics[width=\textwidth]{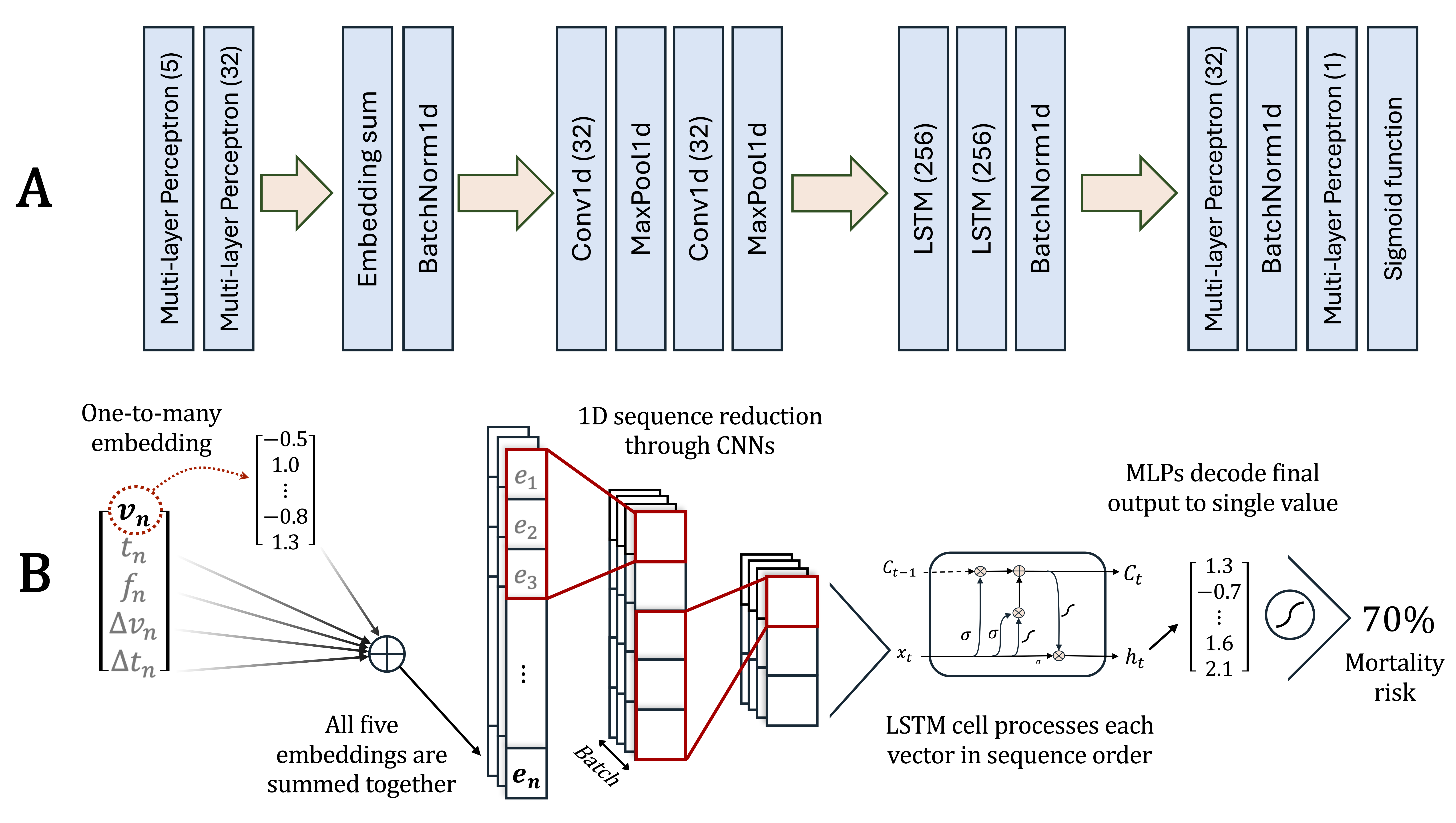}
    \caption{A) CNN-LSTM Model Architecture. B) Visualisation of how data flows through the model.}
    \label{fig:model-architecture}
\end{figure*}

\subsection{Data Selection and Pre-Processing}
\subsubsection{MIMIC-IV}
The Medical Information Mart for Intensive Care IV (MIMIC-IV v3.0) database is a collection of de-identified electronic healthcare data for more than 360,000 patients admitted to Beth Israel Deaconess Medical Center, USA between 2008 - 2022 \citep{mimic-1, mimic-2}. Our training data consisted of 65,000 patients across 85,000 hospital admissions with at least one of the required input features. These included a mixture of elective post-operative and emergency surgical/medical admissions. The input features consisted of a range of biomarkers, intravenous medications, and demographic information (Appendix \ref{ap:features}). The patients were divided into train (80\%), validation (10\%), and test (10\%) groups. All input data were standardised based on mean and standard deviation values computed from the training group. Additional per-feature processing steps are described in Appendix \ref{ap:pre-processing}.
\subsubsection{SICDB}
For external validation, we employed the Salzburg Intensive Care database (SICdb v1.0.6), a publicly available European dataset containing de-identified healthcare data for more than 27,000 ICU admissions across 21,000 patients admitted to the University Hospital Salzburg between 2013 and 2021 \citep{sicdb-1, sicdb-2}. The data were standardised using the mean and standard deviation values computed from the MIMIC-IV training group.

\subsection{Data Representation and State Definition} \label{subsection:state-definition}
\subsubsection{Time Series as 1D Sequences} \label{subsubsection:measurement-tuples}
Each patient's time series data was represented as a 1D sequence of tuples \(\{v, t, f, \Delta v, \Delta t\} \). Here, \(v\) represents the measurement value, \(t\) the time-point relative to the current state marker (\ref{subsubsection:state-markers}), and \(f\) the feature label. Additionally, \(\Delta v\) denotes the change in value (if available) since the last measurement of the same feature, and \(\Delta t\) indicates the time interval between these measurements.

\subsubsection{State Markers and Input Data Construction}
\label{subsubsection:state-markers}
In our framework, each measurement in a patient's admission is treated as a unique ``state marker", from which the model can learn the patient's latent state at that point in time. Specifically, the model is provided with the observation data  \(\mathcal{O}_t\), which consists of the state marker and up to 396 retrospective measurements taken over the previous 7 days  (plus age, gender, and weight). The model processes the data as an ordered sequence of real measurements (\ref{subsubsection:measurement-tuples} and Figure \hyperlink{\figureref{fig:model-architecture}}{2B}), with the temporal component encoded within the tuple rather than the sequence position itself. The chosen model architecture (\ref{subsubsection:model-architecture}) can accommodate sequences of variable lengths, and thus there is almost no missing data or requirement for data imputation (with the exception of patient weight, described in Appendix \ref{ap:pre-processing}).

\begin{table*}[t]
\begin{center}
\small
\begin{tabular}{|p{5cm}p{3cm}p{3cm}|}
\hline
\textbf{Descriptor} & \textbf{MIMIC-IV} & \textbf{SICdb} \\
\hline
Unique patients & 65,050 & 21,447 \\
Unique hospital admissions & 84,659 & \textit{not available} \\
Unique ICU admissions & 93,508 & 27,213 \\
ICU length-of-stay, mean days (std) & 3.7 (5.4) & 3.5 (6.4) \\
&&\\
Hospital mortality\(^*\) (\%) & 11,126 (11.9\%) & 2,127 (7.8\%) \\
1-year mortality (\%) & 23,655 (27.9\%) & 5,071 (18.6\%) \\
&&\\
Median age (IQR) & 66 (55 - 77) & 70 (60 - 75) \\
Female (\%) & 28,496 (43.8\%) & 8,170 (38.1\%) \\
&&\\
\footnotesize \textit{\(^*\)Relative to each ICU admission} &&\\
\hline
\end{tabular}
\caption{Baseline characteristics of the datasets.}
\label{tab:dataset-descriptions}
\end{center}
\end{table*}

\subsubsection{Choice of Next State}
When determining our distribution \(\mathcal{D}\), it is important that our step sizes are large enough to demonstrate a clear trajectory, but not so large as to suffer excessive variance. We defined an eligibility window of 24 hours, with the start of the window delayed \(x\) hours into the future. The ``next state" marker is chosen as the first available measurement occurring inside this window - if no measurements are made within the eligible period, we assume the patient has reached a terminal state (i.e., the current state marker was measured in the final 24 hours before discharge or death). We experimented with a range of possible delays when training the TD model, with \(x\) set as either 4, 16, 24, 48, 72, or 120 hours. The best-performing model (on the validation dataset) was then chosen for further analysis.

\subsection{Models} \label{subsection:models}
\subsubsection{Model Architecture}
\label{subsubsection:model-architecture}

The model architecture is demonstrated in Figure \hyperlink{\figureref{fig:model-architecture}}{2A}. The input data consists of a sequence of observed measurements, in the form of a measurement tuple \(\{v, t, f, \Delta v, \Delta t\}\) (\ref{subsubsection:measurement-tuples}). Each tuple is embedded using five one-to-many multi-layer perceptron (MLP) networks, corresponding to each component of the tuple, before summing the embeddings together to create an embedded measurement $e_n$ \citep{self-supervised-health}. The sequence of embedded measurements is then processed through CNN (for sequence length reduction) and LSTM (for sequence processing) layers, a popular machine learning architecture which has the advantage of being able to process sequences of variable length. The final hidden state of the terminal LSTM is decoded by two densely connected MLPs to give a single output for mortality risk. ReLU activation is used in all hidden layers, with sigmoid activation for the final output layer.

\subsubsection{Candidate Models}
\label{subsubsection:candidate-models}
We trained six groups of models using the same base architecture but different learning targets: our TD model, and five baseline models trained with supervised learning. The TD model was trained according to Equation \eqref{eq:mortality-prediction-td-learning}, with the terminal reward set to 28-day mortality in the terminal state. The target for the supervised models was each state's observed mortality at one of several pre-defined time horizons (1 day, 3 days, 7 days, 14 days, and 28 days). This allows for a comparative evaluation of TD model performance across a spectrum of well-defined temporal horizons.

\subsubsection{Model Training}
\label{subsubsection:model-training}
Each candidate model was independently trained and evaluated five times. We used a binary cross-entropy loss, but included an optional class balancing factor for the supervised baselines (in which the loss is weighted according to the normalised inverse class frequency). This can often optimise supervised learning performance when training on class-imbalanced datasets such as MIMIC-IV.

Using the same network for both the predicted and target values can lead to training instability. To address this, we implement a separate identical ``target network" that provides stable targets for the main network, and is gradually synchronised with the main network after each update \citep{soft-updates-q-learning}. More details for this and other training hyperparameters can be found in Appendix \ref{ap:training-hyperparameters}.

\subsubsection{Model Evaluation}
\label{subsubsection:model-evaluation}

The discriminative ability of all models was  evaluated on the internal and external test data using the area under the receiver operating characteristic curve (AUROC) for each of the possible mortality labels – i.e., how well does the model's predicted score discriminate between classes when the label is set to mortality at 1 day, 3 days, 7 days, 14 days, or 28 days. AUROC scores were also calculated using the Sequential Organ Failure Assessment (SOFA) score for samples in MIMIC-IV to provide a clinical baseline.

\subsection{Statistical Testing}
The mean performance of the TD model was compared to the mean performance of each baseline model in each evaluation using a one-tailed paired Student's t-test, with Benjamini-Yekutieli correction to account for multiple testing under dependency \citep{benjamini-yekutieli}. No comparison was made between the baseline models themselves.

\subsection{Software}
All training with performed using Python 3.11. Models were custom-built using PyTorch 2.3.1, and evaluated using the TorchEval 0.0.7 evaluation and SciPy 1.13 statistical testing packages.

\section{Results}
\label{sec:results}
\subsection{Dataset Baseline Characteristics}

The baseline characteristics of the two datasets are summarised in Table \ref{tab:dataset-descriptions}. Both datasets exhibit similar ICU lengths of stay, with comparable means and standard deviations in the number of days in ICU. The patient cohorts in both datasets share similar median ages and gender distributions, which suggests a level of comparability in terms of the general patient population. 

\begin{figure}[htpb]
\floatconts
  {fig:state-interval-performance}
  {\caption{The effect of the size of interval between states during training on TD model performance (on the validation dataset). The chosen model, TD-24hr, is marked with an asterisk.}}
  {\includegraphics[width=0.9\linewidth]{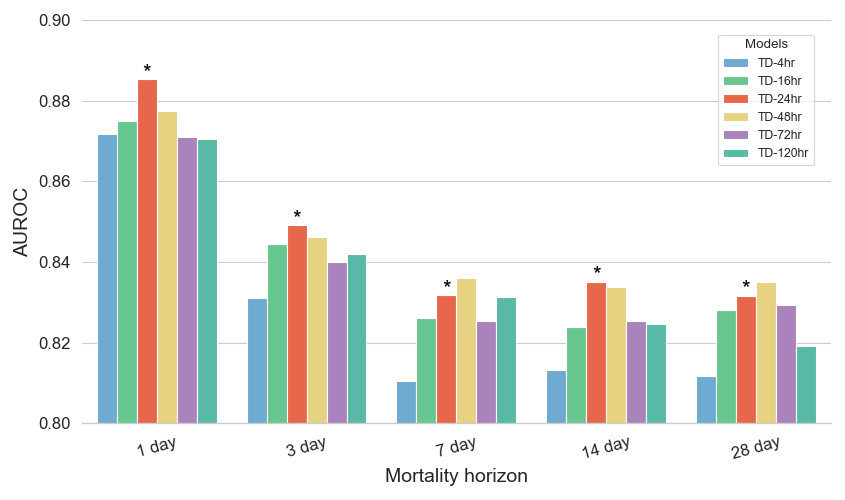}}
\end{figure}

However, there is a notable difference in mortality outcomes, with the MIMIC-IV dataset showing a significantly higher average hospital and 1-year mortality rate. This may relate to differences in underlying co-morbidities and admission diagnoses for the two populations, as well as differences in the collection period.

\begin{figure}[htbp]
\floatconts
  {fig:time-distribution}
  {\caption{The distribution of time between ``states" and ``next states".}}
  {\includegraphics[width=0.9\linewidth]{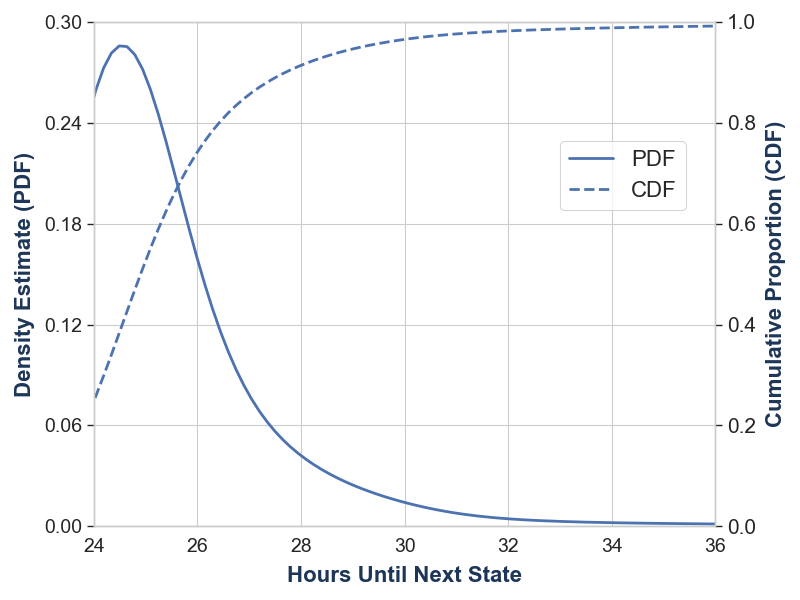}}
\end{figure}

\subsection{Effect of State Interval Size on TD Performance}
The impact on TD model performance when trained with different sized delays between states is shown in \figureref{fig:state-interval-performance}. The overall best-performing model was trained with a state-to-state delay of 24 hours (although a longer delay of 48 hours performed better for 7-day and 28-day mortality prediction). We thus chose the TD-24hr model as our benchmark model for subsequent analyses. For a 24-hour delay between states, the distribution of exact intervals between ``states" and ``next states" can be seen in \figureref{fig:time-distribution}, with 50\% of ``next state" markers occurring within 1 hour of the eligible period, and 90\% occurring within 4 hours.

\subsection{Model Evaluation on the Internal Dataset} \label{subsection:internal-results}

\begin{figure*}[t]
\floatconts
  {fig:all-results-balanced}
  {\caption{A) AUROC evaluation scores for the TD and supervised models\(^+\) on different mortality horizons on the internal MIMIC-IV dataset. B) Evaluation scores for the same models on the external SICdb dataset. \newline\newline \footnotesize \textit{\(^*\)p\(\leq\)0.05, \(^{**}\)p\(\leq\)0.01, \(^{***}\)p\(\leq\)0.001} \newline \(^+\)\textit{only those trained with balanced cross-entropy shown.}}}
  {\includegraphics[scale=0.5]{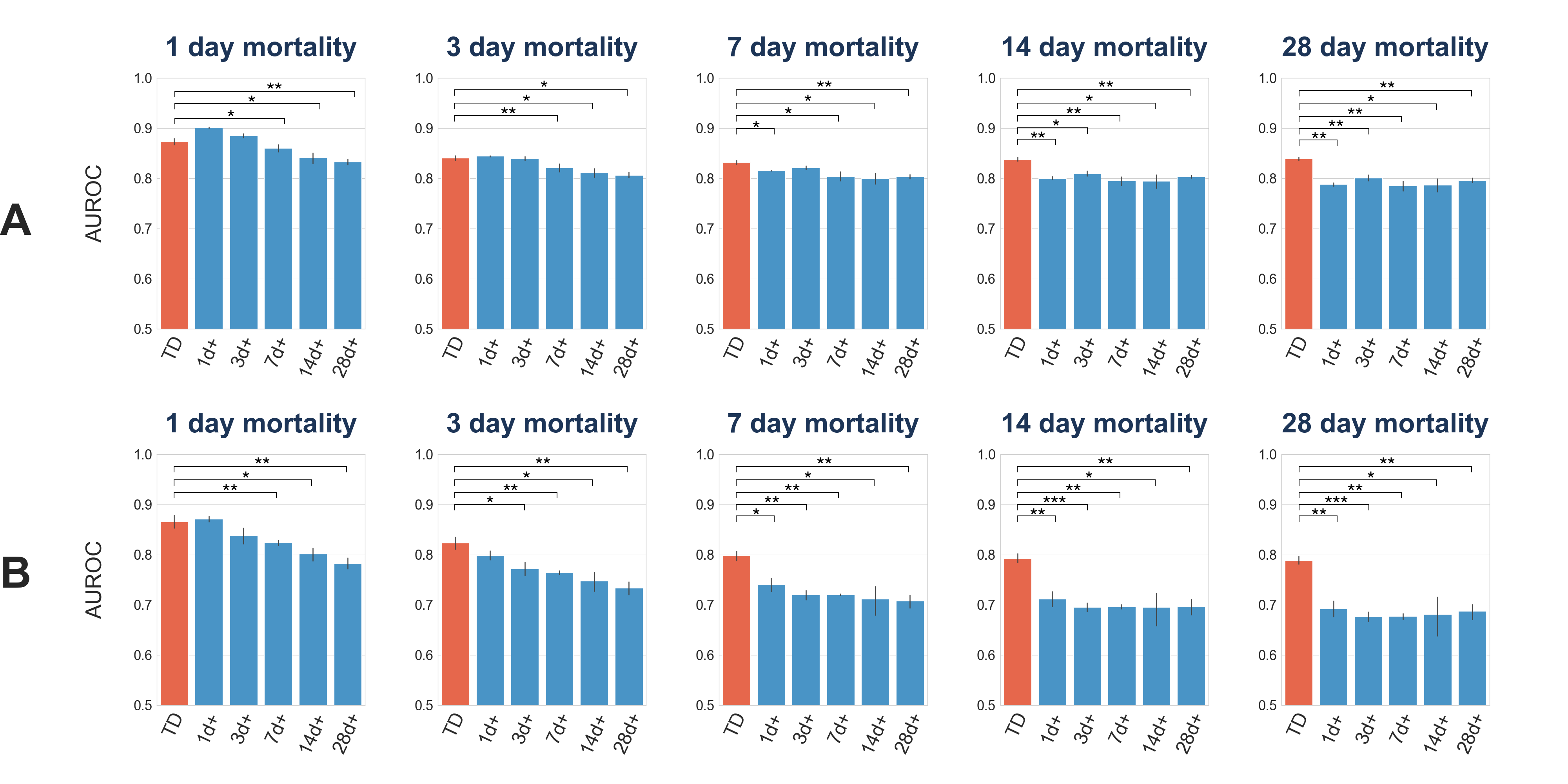}}
\end{figure*}

The evaluation results for each model group\footnote{In nearly all cases, the supervised models trained with class weighting outperformed those without - full results are reported in Appendix \ref{ap:additional-results}.} on the MIMIC-IV dataset are summarised in Table \ref{tab:internal-dataset-auroc} and Figure \hyperlink{\figureref{fig:all-results-balanced}}{5A}. When predicting short-term mortality, the models trained via supervised learning on short-term mortality labels achieved the highest performance, with AUROC scores as high as 0.90. The predictions of the short-term models also had high AUROC scores for longer-term mortality, frequently outperforming the supervised models trained specifically on long-term labels. However, the TD model was found to consistently exceed the results of \textit{all} baseline models when evaluated on longer time horizons (7 to 28 days). All models significantly outperformed the clinical SOFA score (0.69-0.74), which had AUROC scores consistent with other publications \citep{sofa-auroc-1, sofa-auroc-2, sofa-auroc-3}.

\subsection{External Model Validation} \label{subsection:external-results}
Each model was evaluated externally on the SICdb dataset, with results presented in Table \ref{tab:external-dataset-auroc} and Figure \hyperlink{\figureref{fig:all-results-balanced}}{5B}. All models exhibited some degree of performance degradation on the unseen data, and this decline is most pronounced when predicting long-term mortality. When compared to the supervised models, the TD model showed noticeably less deterioration in predictive performance over the extended time horizons, and consistently outperformed all supervised models in all mortality predictions beyond the ultra-short (1-day) horizon.

\begin{table}[ht]
\begin{center}
\tabcolsep=3.0pt\relax
\small
\tablefirsthead{
    \multicolumn{2}{c}{} & \multicolumn{5}{c}{Models} \\
    Mortality & SOFA & \textit{1d\(^+\)} & \textit{3d\(^+\)} & \textit{7d\(^+\)} & \textit{14d\(^+\)} & \textit{28d\(^+\)} & \multicolumn{1}{c|}{\textit{TD}}\\}
\tablelasttail{
\hline
}
\bottomcaption{Mean AUROC results on the internal MIMIC-IV dataset.
\newline\(^+\)\textit{indicates training with balanced cross-entropy.}}
\label{tab:internal-dataset-auroc}
\begin{supertabular}
{|c||c|ccccc|c|}
\hline
1-day & 0.746 & \textbf{0.901} & 0.885 & 0.860 & 0.842 & 0.833 & 0.874 \\
3-day & 0.718 &\textbf{0.844} & 0.840 & 0.821 & 0.811 & 0.807 & 0.841 \\
7-day & 0.700 &0.816 & 0.821 & 0.804 & 0.800 & 0.803 & \textbf{0.832} \\
14-day & 0.697 & 0.800 & 0.809 & 0.795 & 0.795 & 0.803 & \textbf{0.838} \\
28-day & 0.694 & 0.788 & 0.801 & 0.785 & 0.786 & 0.796 & \textbf{0.839} \\
\hline
\end{supertabular}
\end{center}
\end{table}

\subsection{Statistical Analysis}

Each model was trained five times, and the resulting mean and standard deviation was used to construct 95\% confidence intervals, which are demonstrated in \figureref{fig:all-results-balanced}. Most models were found to have relatively tight 95\% CI bounds for their internal performance, with these bounds typically widening when evaluated on the external dataset. Student's t-test values were also calculated, with adjustments to account for multiple testing. The TD model outperformed the baseline supervised models to high degrees of statistical significance (\(p\leq0.01\)), with the significance of these results increasing for longer prediction horizons and during external evaluation.

\begin{table}[t]
\begin{center}
\tabcolsep=3.0pt\relax
\small
\begin{tabular}{|c||ccccc|c|}
\multicolumn{1}{c}{} & \multicolumn{5}{c}{Models} & \multicolumn{1}{c}{} \\
Mortality & \textit{1d\(^+\)} & \textit{3d\(^+\)} & \textit{7d\(^+\)} & \textit{14d\(^+\)} & \textit{28d\(^+\)} & \textit{TD} \\
\hline
1 day & \textbf{0.871} & 0.838 & 0.824 & 0.801 & 0.783 & 0.865 \\
3 days & 0.799 & 0.772 & 0.765 & 0.748 & 0.734 & \textbf{0.823} \\
7 days & 0.741 & 0.720 & 0.720 & 0.712 & 0.708 & \textbf{0.797}\\
14 days & 0.712 & 0.695 & 0.696 & 0.695 & 0.697 & \textbf{0.792}\\
28 days & 0.692 & 0.676 & 0.677 & 0.681 & 0.688 & \textbf{0.788}\\
\hline
\end{tabular}
\end{center}
\caption{Mean AUROC results on the external SICdb dataset.\newline \(^+\)\textit{ indicates training with balanced cross-entropy.}}
\label{tab:external-dataset-auroc}
\end{table}

\section{Discussion}
\label{sec:discuss}
In this report, we have demonstrated that models trained with temporal difference learning are able to produce long-term mortality predictions superior to those derived from conventional supervised learning (and far superior to clinical scores such as SOFA). This makes intuitive sense - as mentioned earlier, ICU patients can have lengths of stay that range from a matter of hours to many months, and an admission with a long duration can be expected to have greater variance for its final outcome. As machine learning algorithms become more complex, these models are then at risk of over-fitting to noise for these distant labels, and may deteriorate significantly when validated externally. We observe this in our results, with the supervised models trained directly on longer-term labels tending to perform worse on both the internal and external evaluations.

On the other hand, models trained on identical input data using temporal difference learning can learn more accurate predictions for long-term outcomes by generalising to the near-term trajectories of each state. This shortens the horizon of the target label to just a single step in the future, reducing the variance and degree of over-fitting, and does not appear to suffer significant bias for a sufficiently sized training dataset. Importantly, the prediction accuracy remains robust even when validated on an external dataset collected from another continent. When combined with our choice of model architecture, this produces a model that generalises well; can be applied at any stage in an ICU patient's admission; can process any length of inpatient stay; and does not suffer from data missingness.

From a purely theoretical standpoint, there are several reasons why TD learning could fail when applied to healthcare settings. First, the states are only partially observable, and it may not be possible for the model to infer an accurate latent state \(\mathcal{S}_t\) for the patient given a limited set of visible observations $\mathcal{O}_t$. Second, the irregular distribution of time intervals (\figureref{fig:time-distribution}) could limit the model's ability to consistently infer transitions between states. Third, two patients will never truly occupy the same continuous state space, and the model may fail to group ``similar" patient observations together as part of trajectory mixing. Fourth, patient states may not observe a consistent transition probability over time, a key assumption of the SMRP. Despite all of these potential limitations, we have managed to show experimentally that models trained using TD learning are still able to converge to a coherent and accurate prediction for the complex ICU patient.

There remain several avenues for further research. Ideally, TD learning should be evaluated on multiple architectures (e.g., Transformers, RNNs), with a sensitivity analysis to assess the impact of training dataset size. For instance, how does the potential for bias in TD learning evolve with smaller datasets, and at what sizes might it become significantly problematic? We aim to explore these questions in future work.

\section{Conclusions}
\label{sec:conclusion}
This report proposes a framework for applying temporal difference learning to outcome prediction tasks when using irregular time series health data. We subsequently implement this framework for the task of mortality estimation in the ICU, chosen for its partially observable data and long, complex trajectories. When compared to standard supervised learning methods, models trained with temporal difference learning can be shown to predict distant health outcomes with a higher level of accuracy, and show greater resilience when evaluated on external unseen datasets. This work has important implications both for the implementation of temporal difference learning within healthcare reinforcement learning, and also for the wider field of health risk estimation.

\acks{We gratefully acknowledge the funding and facilities provided by the UKRI Centre for Doctoral Training in AI-enabled Healthcare (grant EP/S021612/1) and University College London respectively, which enabled this research. The views expressed in the text are those of the authors and do not necessarily reflect those of the above bodies.}

\section*{Orcid}
\textit{Thomas Frost} \orcidlink{https://orcid.org/0009-0002-5990-5800} \href{https://orcid.org/0009-0002-5990-5800}{https://orcid.org/0009-0002-5990-5800}
\textit{Ken Li} \orcidlink{https://orcid.org/0000-0003-3073-3128} \href{https://orcid.org/0000-0003-3073-3128}{https://orcid.org/0000-0003-3073-3128}
\textit{Steve Harris} \orcidlink{https://orcid.org/0000-0002-4982-1374} \href{https://orcid.org/0000-0002-4982-1374}{https://orcid.org/0000-0002-4982-1374}

\bibliography{references}

\begin{thebibliography}{30}
\providecommand{\natexlab}[1]{#1}
\providecommand{\url}[1]{\texttt{#1}}
\expandafter\ifx\csname urlstyle\endcsname\relax
  \providecommand{\doi}[1]{doi: #1}\else
  \providecommand{\doi}{doi: \begingroup \urlstyle{rm}\Url}\fi

\bibitem[Benjamini and Yekutieli(2001)]{benjamini-yekutieli}
Yoav Benjamini and Daniel Yekutieli.
\newblock The control of the false discovery rate in multiple testing under
  dependency.
\newblock \emph{Annals of statistics}, pages 1165--1188, 2001.

\bibitem[Capuzzo et~al.(2014)Capuzzo, Volta, Tassinati, Moreno, Valentin,
  Guidet, Iapichino, Martin, Perneger, Combescure,
  et~al.]{icu-mortality-europe}
Maurizia Capuzzo, Carlo~Alberto Volta, Tania Tassinati, Rui~Paulo Moreno,
  Andreas Valentin, Bertrand Guidet, Gaetano Iapichino, Claude Martin, Thomas
  Perneger, Christophe Combescure, et~al.
\newblock Hospital mortality of adults admitted to intensive care units in
  hospitals with and without intermediate care units: a multicentre european
  cohort study.
\newblock \emph{Critical Care}, 18:\penalty0 1--15, 2014.

\bibitem[Checkley et~al.(2014)Checkley, Martin, Brown, Chang, Dabbagh, Fremont,
  Girard, Rice, Howell, Johnson, et~al.]{icu-mortality-usa}
William Checkley, Greg~S Martin, Samuel~M Brown, Steven~Y Chang, Ousama
  Dabbagh, Richard~D Fremont, Timothy~D Girard, Todd~W Rice, Michael~D Howell,
  Steven~B Johnson, et~al.
\newblock Structure, process, and annual icu mortality across 69 centers:
  United states critical illness and injury trials group critical illness
  outcomes study.
\newblock \emph{Critical care medicine}, 42\penalty0 (2):\penalty0 344--356,
  2014.

\bibitem[Esmaeili~Tarki et~al.(2023)Esmaeili~Tarki, Afaghi, Rahimi, Kiani,
  Varahram, and Abedini]{sofa-auroc-2}
Farzad Esmaeili~Tarki, Siamak Afaghi, Fatemeh~Sadat Rahimi, Arda Kiani,
  Mohammad Varahram, and Atefeh Abedini.
\newblock Serial sofa-score trends in icu-admitted covid-19 patients as
  predictor of 28-day mortality: A prospective cohort study.
\newblock \emph{Health Science Reports}, 6\penalty0 (5):\penalty0 e1116, 2023.

\bibitem[Garc{\'\i}a-Gallo et~al.(2020)Garc{\'\i}a-Gallo, Fonseca-Ruiz, Celi,
  and Duitama-Mu{\~n}oz]{ml-icu-mortality-no-ext-1}
JE~Garc{\'\i}a-Gallo, NJ~Fonseca-Ruiz, LA~Celi, and JF~Duitama-Mu{\~n}oz.
\newblock A machine learning-based model for 1-year mortality prediction in
  patients admitted to an intensive care unit with a diagnosis of sepsis.
\newblock \emph{Medicina intensiva}, 44\penalty0 (3):\penalty0 160--170, 2020.

\bibitem[Goldberger et~al.(2000)Goldberger, Amaral, Glass, Hausdorff, Ivanov,
  Mark, Mietus, Moody, Peng, and Stanley]{physionet}
Ary~L Goldberger, Luis~AN Amaral, Leon Glass, Jeffrey~M Hausdorff, Plamen~Ch
  Ivanov, Roger~G Mark, Joseph~E Mietus, George~B Moody, Chung-Kang Peng, and
  H~Eugene Stanley.
\newblock Physiobank, physiotoolkit, and physionet: components of a new
  research resource for complex physiologic signals.
\newblock \emph{circulation}, 101\penalty0 (23):\penalty0 e215--e220, 2000.

\bibitem[Iwase et~al.(2022)Iwase, Nakada, Shimada, Oami, Shimazui, Takahashi,
  Yamabe, Yamao, and Kawakami]{ml-icu-mortality-no-ext-3}
Shinya Iwase, Taka-aki Nakada, Tadanaga Shimada, Takehiko Oami, Takashi
  Shimazui, Nozomi Takahashi, Jun Yamabe, Yasuo Yamao, and Eiryo Kawakami.
\newblock Prediction algorithm for icu mortality and length of stay using
  machine learning.
\newblock \emph{Scientific reports}, 12\penalty0 (1):\penalty0 12912, 2022.

\bibitem[Johnson et~al.(2020)Johnson, Bulgarelli, Pollard, Horng, Celi, and
  Mark]{mimic-1}
Alistair Johnson, Lucas Bulgarelli, Tom Pollard, Steven Horng, Leo~Anthony
  Celi, and Roger Mark.
\newblock Mimic-iv.
\newblock \emph{PhysioNet. Available online at: https://physionet.
  org/content/mimiciv/1.0/(accessed August 23, 2021)}, pages 49--55, 2020.

\bibitem[Johnson et~al.(2023)Johnson, Bulgarelli, Shen, Gayles, Shammout,
  Horng, Pollard, Hao, Moody, Gow, et~al.]{mimic-2}
Alistair~EW Johnson, Lucas Bulgarelli, Lu~Shen, Alvin Gayles, Ayad Shammout,
  Steven Horng, Tom~J Pollard, Sicheng Hao, Benjamin Moody, Brian Gow, et~al.
\newblock Mimic-iv, a freely accessible electronic health record dataset.
\newblock \emph{Scientific data}, 10\penalty0 (1):\penalty0 1, 2023.

\bibitem[Kim et~al.(2019)Kim, Kim, Cho, Kim, Sol, Sung, Cho, Park, Jang, Kim,
  et~al.]{real-time-icu-mortality-with-ext-2}
Soo~Yeon Kim, Saehoon Kim, Joongbum Cho, Young~Suh Kim, In~Suk Sol, Youngchul
  Sung, Inhyeok Cho, Minseop Park, Haerin Jang, Yoon~Hee Kim, et~al.
\newblock A deep learning model for real-time mortality prediction in
  critically ill children.
\newblock \emph{Critical care}, 23:\penalty0 1--10, 2019.

\bibitem[Kim et~al.(2021)Kim, Ausin, and Chi]{td-learning-continuous-time}
Yeo~Jin Kim, Markel~Sanz Ausin, and Min Chi.
\newblock Multi-temporal abstraction with time-aware deep q-learning for septic
  shock prevention.
\newblock In \emph{2021 IEEE International Conference on Big Data (Big Data)},
  pages 1657--1663. IEEE, 2021.

\bibitem[Ko et~al.(2018)Ko, Shim, Lee, Kim, and Yoon]{apache-performance}
Mihye Ko, Miyoung Shim, Sang-Min Lee, Yujin Kim, and Soyoung Yoon.
\newblock Performance of apache iv in medical intensive care unit patients:
  comparisons with apache ii, saps 3, and mpm0 iii.
\newblock \emph{Acute and critical care}, 33\penalty0 (4):\penalty0 216--221,
  2018.

\bibitem[Komorowski et~al.(2018)Komorowski, Celi, Badawi, Gordon, and
  Faisal]{td-learning-healthcare-1}
Matthieu Komorowski, Leo~A Celi, Omar Badawi, Anthony~C Gordon, and A~Aldo
  Faisal.
\newblock The artificial intelligence clinician learns optimal treatment
  strategies for sepsis in intensive care.
\newblock \emph{Nature medicine}, 24\penalty0 (11):\penalty0 1716--1720, 2018.

\bibitem[Lei et~al.(2023)Lei, Han, Wang, Han, Fang, Lin, and
  Huang]{ml-icu-mortality-with-ext-1}
Mingxing Lei, Zhencan Han, Shengjie Wang, Tao Han, Shenyun Fang, Feng Lin, and
  Tianlong Huang.
\newblock A machine learning-based prediction model for in-hospital mortality
  among critically ill patients with hip fracture: An internal and external
  validated study.
\newblock \emph{Injury}, 54\penalty0 (2):\penalty0 636--644, 2023.

\bibitem[Levine et~al.(2020)Levine, Kumar, Tucker, and Fu]{levine2020offline}
Sergey Levine, Aviral Kumar, George Tucker, and Justin Fu.
\newblock Offline reinforcement learning: Tutorial, review, and perspectives on
  open problems.
\newblock \emph{arXiv preprint arXiv:2005.01643}, 2020.

\bibitem[Lillicrap(2015)]{soft-updates-q-learning}
TP~Lillicrap.
\newblock Continuous control with deep reinforcement learning.
\newblock \emph{arXiv preprint arXiv:1509.02971}, 2015.

\bibitem[Liu et~al.(2024)Liu, Xie, Shu, Chen, Sun, Zhong, Liang, Li, Yang, Han,
  et~al.]{td-learning-healthcare-3}
Jiang Liu, Yihao Xie, Xin Shu, Yuwen Chen, Yizhu Sun, Kunhua Zhong, Hao Liang,
  Yujie Li, Chunyong Yang, Yan Han, et~al.
\newblock Value function assessment to different rl algorithms for heparin
  treatment policy of patients with sepsis in icu.
\newblock \emph{Artificial Intelligence in Medicine}, 147:\penalty0 102726,
  2024.

\bibitem[Mandrekar(2010)]{auroc-curves}
Jayawant~N Mandrekar.
\newblock Receiver operating characteristic curve in diagnostic test
  assessment.
\newblock \emph{Journal of Thoracic Oncology}, 5\penalty0 (9):\penalty0
  1315--1316, 2010.

\bibitem[Meyer et~al.(2018)Meyer, Zverinski, Pfahringer, Kempfert, Kuehne,
  S{\"u}ndermann, Stamm, Hofmann, Falk, and
  Eickhoff]{real-time-icu-mortality-with-ext-1}
Alexander Meyer, Dina Zverinski, Boris Pfahringer, J{\"o}rg Kempfert, Titus
  Kuehne, Simon~H S{\"u}ndermann, Christof Stamm, Thomas Hofmann, Volkmar Falk,
  and Carsten Eickhoff.
\newblock Machine learning for real-time prediction of complications in
  critical care: a retrospective study.
\newblock \emph{The Lancet Respiratory Medicine}, 6\penalty0 (12):\penalty0
  905--914, 2018.

\bibitem[Nistal-Nu{\~n}o(2022)]{ml-icu-mortality-no-ext-2}
Beatriz Nistal-Nu{\~n}o.
\newblock Developing machine learning models for prediction of mortality in the
  medical intensive care unit.
\newblock \emph{Computer Methods and Programs in Biomedicine}, 216:\penalty0
  106663, 2022.

\bibitem[Prasad et~al.(2017)Prasad, Cheng, Chivers, Draugelis, and
  Engelhardt]{td-learning-healthcare-2}
Niranjani Prasad, Li-Fang Cheng, Corey Chivers, Michael Draugelis, and
  Barbara~E Engelhardt.
\newblock A reinforcement learning approach to weaning of mechanical
  ventilation in intensive care units.
\newblock \emph{arXiv preprint arXiv:1704.06300}, 2017.

\bibitem[Rodemund et~al.(2023{\natexlab{a}})Rodemund, Kokoefer, Wernly, and
  Cozowicz]{sicdb-1}
Niklas Rodemund, Andreas Kokoefer, Bernhard Wernly, and Crispiana Cozowicz.
\newblock Salzburg intensive care database (sicdb), a freely accessible
  intensive care database.
\newblock \emph{PhysioNet https://doi. org/10.13026/ezs8-6v88},
  2023{\natexlab{a}}.

\bibitem[Rodemund et~al.(2023{\natexlab{b}})Rodemund, Wernly, Jung, Cozowicz,
  and Kok{\"o}fer]{sicdb-2}
Niklas Rodemund, Bernhard Wernly, Christian Jung, Crispiana Cozowicz, and
  Andreas Kok{\"o}fer.
\newblock The salzburg intensive care database (sicdb): an openly available
  critical care dataset.
\newblock \emph{Intensive care medicine}, 49\penalty0 (6):\penalty0 700--702,
  2023{\natexlab{b}}.

\bibitem[Sarkar et~al.(2021)Sarkar, Martin, Mattie, Gichoya, Stone, and
  Celi]{sofa-mimic-performance}
Rahuldeb Sarkar, Christopher Martin, Heather Mattie, Judy~Wawira Gichoya,
  David~J Stone, and Leo~Anthony Celi.
\newblock Performance of intensive care unit severity scoring systems across
  different ethnicities in the usa: a retrospective observational study.
\newblock \emph{The Lancet Digital Health}, 3\penalty0 (4):\penalty0
  e241--e249, 2021.

\bibitem[Sutton(2018)]{sutton-barto-book}
Richard~S Sutton.
\newblock Reinforcement learning: An introduction.
\newblock \emph{A Bradford Book}, 2018.

\bibitem[Tipirneni and Reddy(2022)]{self-supervised-health}
Sindhu Tipirneni and Chandan~K Reddy.
\newblock Self-supervised transformer for sparse and irregularly sampled
  multivariate clinical time-series.
\newblock \emph{ACM Transactions on Knowledge Discovery from Data (TKDD)},
  16\penalty0 (6):\penalty0 1--17, 2022.

\bibitem[Wang et~al.(2022{\natexlab{a}})Wang, Wang, Jiang, Du, Zhu, and
  Xi]{sofa-auroc-3}
Na~Wang, Meiping Wang, Li~Jiang, Bin Du, Bo~Zhu, and Xiuming Xi.
\newblock The predictive value of the oxford acute severity of illness score
  for clinical outcomes in patients with acute kidney injury.
\newblock \emph{Renal Failure}, 44\penalty0 (1):\penalty0 320--328,
  2022{\natexlab{a}}.

\bibitem[Wang and Aitchison(2024)]{adamw-weight-decay}
Xi~Wang and Laurence Aitchison.
\newblock How to set adamw's weight decay as you scale model and dataset size.
\newblock \emph{arXiv preprint arXiv:2405.13698}, 2024.

\bibitem[Wang et~al.(2022{\natexlab{b}})Wang, Zhao, and
  Petzold]{td-learning-healthcare-4}
Yuqing Wang, Yun Zhao, and Linda Petzold.
\newblock Predicting the need for blood transfusion in intensive care units
  with reinforcement learning.
\newblock In \emph{Proceedings of the 13th ACM International Conference on
  Bioinformatics, Computational Biology and Health Informatics}, pages 1--10,
  2022{\natexlab{b}}.

\bibitem[Zhou et~al.(2024)Zhou, Lu, Liu, Wang, Zhou, Cui, Zhang, Xiao, Hua,
  Zhu, et~al.]{sofa-auroc-1}
Shu Zhou, Zongqing Lu, Yu~Liu, Minjie Wang, Wuming Zhou, Xuanxuan Cui, Jin
  Zhang, Wenyan Xiao, Tianfeng Hua, Huaqing Zhu, et~al.
\newblock Interpretable machine learning model for early prediction of 28-day
  mortality in icu patients with sepsis-induced coagulopathy: development and
  validation.
\newblock \emph{European Journal of Medical Research}, 29\penalty0
  (1):\penalty0 14, 2024.

\end{thebibliography}

\appendix

\onecolumn
\section{Full Results} \label{ap:additional-results}

\begin{table}[htbp]
\begin{center}
\tabcolsep=3.0pt\relax
\small
\tablefirsthead{
    \multicolumn{2}{c}{} & \multicolumn{10}{c}{Models} \\
    Mortality & SOFA & \textit{1d} &\textit{1d\(^+\)} & \textit{3d} & \textit{3d\(^+\)} & \textit{7d} & \textit{7d\(^+\)} & \textit{14d} &\textit{14d\(^+\)} & \textit{28d} & \textit{28d\(^+\)} & \multicolumn{1}{c|}{\textit{TD}}\\}
\tablelasttail{
\hline
}
\bottomcaption{Mean AUROC results on the internal MIMIC-IV dataset (all models). \newline\(^+\)\textit{indicates training with balanced cross-entropy.}}
\label{tab:full-results-internal-dataset}
\begin{supertabular}
{|c||c|cccccccccc|c|}
\hline
1-day & 0.746 & 0.898 & \textbf{0.901} & 0.877 & 0.885 & 0.8521 & 0.860 & 0.831 & 0.842 & 0.827 & 0.833 & 0.874 \\
3-days & 0.718 & 0.838 & \textbf{0.844} & 0.831 & 0.840 & 0.814 & 0.821 & 0.799 & 0.811 & 0.800 & 0.807 & 0.841 \\
7-days & 0.700 & 0.806 & 0.816 & 0.813 & 0.821 & 0.802 & 0.804 & 0.787 & 0.800 & 0.798 & 0.803 & \textbf{0.832} \\
14-days & 0.697 & 0.789 & 0.800 & 0.802 & 0.809 & 0.796 & 0.795 & 0.780 & 0.795 & 0.799 & 0.803 & \textbf{0.838} \\
28-days & 0.694 & 0.780 & 0.788 & 0.793 & 0.801 & 0.785 & 0.785 & 0.770 & 0.786 & 0.793 & 0.796 & \textbf{0.839} \\
\hline
\end{supertabular}
\end{center}
\end{table}

\begin{table}[htbp]
\begin{center}
\tabcolsep=3.0pt\relax
\small
\tablefirsthead{
    \multicolumn{1}{c}{} & \multicolumn{10}{c}{Models} \\
    Mortality & \textit{1d} &\textit{1d\(^+\)} & \textit{3d} & \textit{3d\(^+\)} & \textit{7d} & \textit{7d\(^+\)} & \textit{14d} &\textit{14d\(^+\)} & \textit{28d} & \textit{28d\(^+\)} & \multicolumn{1}{c|}{\textit{TD}}\\}
\tablelasttail{
\hline 
}
\bottomcaption{Mean AUROC results on the external SICdb dataset (all models). \newline \(^+\)\textit{indicates training with balanced cross-entropy.}}
\label{tab:full-results-external-dataset}
\begin{supertabular}
{|c||cccccccccc|c|}
\hline
1-day & 0.860 & \textbf{0.871} & 0.839 & 0.838 & 0.810 & 0.824 & 0.789 & 0.801 & 0.776 & 0.783 & 0.865 \\
3-days & 0.782 & 0.799 & 0.776 & 0.772 & 0.755 & 0.765 & 0.732 & 0.748 & 0.728 & 0.734 & \textbf{0.823} \\
7-days & 0.720 & 0.741 & 0.726 & 0.720 & 0.716 & 0.720 & 0.694 & 0.712 & 0.704 & 0.708 & \textbf{0.797}\\
14-days & 0.687 & 0.712 & 0.701 & 0.695 & 0.695 & 0.696 & 0.675 & 0.695 & 
 0.693 & 0.697 & \textbf{0.792}\\
28-days & 0.664 & 0.692 & 0.682 & 0.676 & 0.680 & 0.677 & 0.658 & 0.681 & 0.680 & 0.688 & \textbf{0.788}\\
\hline
\end{supertabular}
\end{center}
\end{table}

\section{Features} \label{ap:features}
\begin{center}
\tablefirsthead{
\hline
\centering \textbf{Category} & \multicolumn{3}{c|}{\textbf{Feature Name}}\\
\hline}
\tablehead{
\hline
\multicolumn{4}{|l|}{\textit{\small continued from previous page}} \\
\hline
\centering \textbf{Category} & \multicolumn{3}{c|}{\textbf{Feature Name}}\\
\hline
}
\tabletail{
\hline
}
\tablelasttail{
\hline
}
\bottomcaption{Input Features}
\label{tab:input-features}
\begin{supertabular}
{|p{4cm}|p{3cm}p{3cm}p{3cm}|}
\textbf{Antiarrhythmics} & Amiodarone &&\\
\hline
\multirow{13}{4em}{\textbf{Antibiotics}} & Aciclovir &  Ambisome & Amikacin \\
& Ampicillin & \multicolumn{2}{l|}{Ampicillin-Sulbactam} \\ 
& Azithromycin & Aztreonam & Caspofungin \\
& Cefazolin & Cefepime & Ceftazidime \\
& Ceftriaxone & Ciprofloxacin & Chloramphenicol \\
& Clindamycin & Colistin & Co-trimoxazole \\
& Daptomycin & Doxycycline & Ertapenem \\
& Erythromycin & Gentamicin & Levofloxacin \\
 & Linezolid & Meropenem & Micafungin \\
& Metronidazole & Nafcillin & Oxacillin \\
& Piperacillin & \multicolumn{2}{l|}{Piperacillin-Tazobactam} \\
 & Rifampin & Tigecycline & Tobramycin \\
& Vancomycin & Voriconazole &\\
\hline
\textbf{Anticoagulants} & \multicolumn{2}{l}{Unfractionated Heparin}&\\
\hline
\textbf{Anticonvulsants} & Levetiracetam & Phenytoin &\\
\hline
\textbf{Antihypertensives} & Nitroglycerin & Nitroprusside &\\
\hline
\textbf{Beta-blockers} & Labetalol &&\\
\hline
\textbf{Blood products} & \multicolumn{2}{l}{Fresh Frozen Plasma} & Red Blood Cells\\
\multirow{2}{12em}{\textbf{Blood products}} & \multicolumn{2}{l}{Human Albumin Solution} & Platelets\\
& \multicolumn{2}{l}{IV Immune Globulin} & \\
\hline
\textbf{Diuretics} & Bumetanide & Furosemide &\\
\hline
\textbf{Glucose control} & Regular Insulin &&\\
\hline
\textbf{Hyperosmotics} & \multicolumn{2}{l}{Hypertonic Saline 3\%} & Mannitol\\
\hline
\textbf{Miscellaneous} & Aminophylline &\multicolumn{2}{l|}{Sodium bicarbonate 8.4\%}\\
\hline
\textbf{Opioids} & Fentanyl & Morphine sulphate\\
\hline
\textbf{Paralytics} & Cisatracurium & Rocuronium & Vecuronium\\
\hline
\multirow{2}{12em}{\textbf{Sedatives}} & Dexmedetomidine & Ketamine & Lorazepam\\
& Midazolam & Propofol & \\
\hline
\textbf{Thrombolytics} & Alteplase & &\\
\hline
\multirow{2}{12em}{\textbf{Vasopressors\textbackslash Inotropes}} & Adrenaline & Dobutamine & Dopamine\\
& Milrinone & Noradrenaline & Vasopressin \\
\hline
\multirow{11}{12em}{\textbf{Laboratory}} & \multicolumn{2}{l}{Alanine transaminase} & Albumin\\
& \multicolumn{2}{l}{Alkaline phosphatase} & Amylase \\
& \multicolumn{2}{l}{Aspartate transferase} & Anion gap \\
& Base excess & Bicarbonate & Blood O\(_2\) pressure \\
& \multicolumn{2}{l}{Blood CO\(_2\) pressure} & Blood pH \\
 & \multicolumn{2}{l}{Blood O\(_2\) saturation} & Calcium (ionised) \\
 & Calcium (total) & C-reactive protein & Chloride\\
  & Creatinine  & Glucose (serum) & Glucose (bedside)\\
 & Haematocrit & Haemoglobin & Lactate \\
  & Lipase & \multicolumn{2}{l|}{Lactate dehydrogenase}\\
  &  Platelet count & Potassium & Prothrombin time \\
 & Sodium & Total bilirubin & Troponin-T\\
  & Urea & \multicolumn{2}{l|}{White blood cell count} \\
\hline
\textbf{Demographics} & Age & Gender & Weight \\
\hline
\end{supertabular}
\end{center}

\section{Additional Data Processing} \label{ap:pre-processing}

In addition to the pre-processing steps described in the methods section, several additional steps were conducted at the feature level. Care was taken to ensure uniform units for all measurements and drug doses across both datasets. IV drugs were separated into bolus events (delivered over $\leq$1 minute) and absolute set rates. For example, a propofol infusion is started at 150mg/hr, with a 20mg bolus at 15 minutes, a rate increase to 200mg/hr after 30 minutes, and then stopped after 2 hours. If we treat the infusion ending as a state marker, this might be recorded as a 150mg/hr rate event at t=120, a 20mg bolus event at t=105, a 200mg/hr rate event at t=90, and a 0mg/hr rate event at t=0. Where two or more continuous infusions were given of the same drug, the rates were combined into a single value for any periods of overlap (i.e., the net rate of drug being delivered). Antibiotic scalar doses were excluded and the administration of an antibiotic was instead treated as a binary ``bolus" event. When generating our model input data, the model can handle cases where there are fewer than 400 available measurements - however, when there were more than 400 measurements available in the eligible window, we prioritised 1) age/gender/weight, followed by 2) all current drug infusion rates, followed by 3) measurements ordered first by novelty (how many times has this feature already occurred), and then by decreasing recency. This latter step ensures that we prioritise including a broad range of features, and then newer information over older information. Outliers were removed according to the 0.005/0.995 quantiles of the training data for drug doses, and 0.001/0.999 quantiles of the training data for laboratory test results (per feature). In the rare cases where patient weight was never recorded, we imputed weight as the mean value for each gender according to the training data (74kg for females, 86kg for males). \(v\) and \(\Delta v\) were standardised to zero mean / unit variance for each feature individually (after outlier removal). \(t\) and \(\Delta t\) were standardised to zero mean / unit variance across all features.

\section{Model Training Hyperparameters} \label{ap:training-hyperparameters}

The AdamW optimiser was used for each backpropagation update. The learning rate and weight decay were set as per \equationref{eq:learning_rate} and \equationref{eq:weight_decay} \citep{adamw-weight-decay}. Each model was trained for a maximum of 10 epochs, with the best iteration selected based on per-epoch evaluation on the validation data. The main and target networks were initialised with identical parameters at the start of training – the target network then receives a soft update of the main network's parameters after every update as per \equationref{eq:target_net_sync}, with $\alpha=0.99$.

\begin{equation}
\textit{Learning rate} = \frac{1}{n_{\textit{trainable parameters}}}
\label{eq:learning_rate}
\end{equation}

\begin{equation}
\textit{Weight decay} = \frac{1}{\textit{learning rate}\cdot n_{\textit{batches per epoch}}}
\label{eq:weight_decay}
\end{equation}

\begin{equation}
\theta_{target} \leftarrow \alpha * \theta_{target} + (1 - \alpha) * \theta_{main}
\label{eq:target_net_sync}
\end{equation}

\end{document}